\documentclass{article}

\usepackage[preprint]{neurips_2026}

\PassOptionsToPackage{numbers,compress}{natbib}

\usepackage[utf8]{inputenc} 
\usepackage[T1]{fontenc}    
\usepackage{hyperref}       
\usepackage{url}            
\usepackage{booktabs}       
\usepackage{amsfonts}       
\usepackage{nicefrac}       
\usepackage{microtype}      
\usepackage{xcolor} 
\usepackage{amsmath}
\usepackage{graphicx}
\usepackage{colortbl}
\usepackage{xcolor}
\definecolor{zbase}{RGB}{31,119,180}
\definecolor{zstar}{RGB}{214,39,40}
\newcommand{\zb}{\textcolor{zbase}{\mathbf{z}}}
\newcommand{\zs}{\textcolor{zstar}{\mathbf{z}^*}}

\title{GVCC: Zero-Shot Video Compression via Codebook-Driven Stochastic Rectified Flow}

\author{%
  Ziyue Zeng\thanks{Equal contribution.}
  \and
  Xun Su\footnotemark[1]
  \and
  Haoyuan Liu
  \and
  Bingyu Lu
  \and
  Yui Tatsumi
  \and
  Hiroshi Watanabe \\
  Graduate School of Fundamental Science and Engineering, Waseda University \\
  Tokyo, Japan \\
  \texttt{zengziyue@fuji.waseda.jp, suxun\_opt@asagi.waseda.jp}
}

\begin{document}

\maketitle

\begin{abstract}
At ultra-low bitrates, high-fidelity reconstruction requires sampling plausible videos from the posterior rather than regressing to oversmoothed conditional means. We propose Generative Video Codebook Codec (GVCC), a zero-shot framework in which a pretrained video generative model serves directly as the decoder, and the transmitted bitstream specifies its generation trajectory. Modern rectified-flow video models are typically sampled with deterministic ODE solvers, which leave no per-step stochastic channel for transmitting compressed information. GVCC addresses this by converting the deterministic flow sampler into an equivalent marginal-preserving stochastic process, so that information can be transmitted by encoding the per-step stochastic innovations. Unlike images, videos introduce longer temporal dependencies and more diverse conditioning modes. We instantiate GVCC in three practical modes: Text-to-Video (T2V) without a reference frame, autoregressive Image-to-Video (I2V) with tail latent correction, and First-Last-Frame-to-Video (FLF2V) with boundary-sharing Group of Pictures (GOP) chaining. On the seven-sequence UVG dataset, local atom-count sweeps characterize the rate--quality behavior of all three variants. We report full-dataset perceptual and fidelity metrics together with temporal diagnostics, without inferring matched-rate or global RD improvements from these limited local sweeps.
\end{abstract}

\section{Introduction}\label{sec:intro}

Video compression at extremely low bitrates remains a long-standing challenge~\cite{lu2019dvc, mentzer2022neural, li2024neural, khan2025perceptual}.
As bitrate decreases, both conventional hybrid codecs and learned neural codecs tend to suffer from severe detail loss and oversmoothing, a manifestation of the rate--distortion--perception trade-off~\cite{blau2019rdp}.
Recent advances in generative modeling have opened a promising direction for perceptual reconstruction.
However, most existing generative video compression methods do not make generation the central mechanism of compression.
Instead, they adopt a hybrid design in which a conventional codec first produces a compressed representation, and a generative model is then applied as a refinement or restoration module.
In such systems, the generative model may improve perceptual quality, but it does not determine the transmitted symbols or the decoder's reconstruction trajectory.

In the image domain, generative compression has evolved from GAN-based learned
decoders~\cite{goodfellow2020generative,esser2021taming,agustsson2019extreme,mentzer2020hific}
to diffusion-based generative priors~\cite{ho2020ddpm,rombach2022high}. Early diffusion codecs such as DiffC~\cite{theis2022diffc} transmit noisy observations and reconstruct images with a generative prior, while conditional diffusion decoders further improve rate--perception trade-offs~\cite{yang2023cdc,hoogeboom2023high,careil2024perco}. More recent codecs extend this paradigm to large pretrained image diffusion models~\cite{relic2024foundation,vonderfecht2025lossy,jia2026coddiffusionfoundationmodel}. At the same time, the generative backbones used by frontier image and video models have shifted from classical diffusion parameterizations toward
rectified-flow formulations~\cite{liu2022rectified,esser2024sd3}, which are now standard in modern video generators. Recent video compression methods have begun to incorporate pretrained video generators~\cite{blattmann2023svd,yang2024cogvideox,wan2025report} for
sequence-level latent refinement~\cite{mao2025gnvcvd}, improving temporal consistency. Nevertheless, these approaches still follow a hybrid paradigm: a separate latent codec defines the compressed representation, while the video generator acts as a post-hoc restorer rather than as the codec itself.

\begin{figure}[t]
\centering
\includegraphics[width=\textwidth]{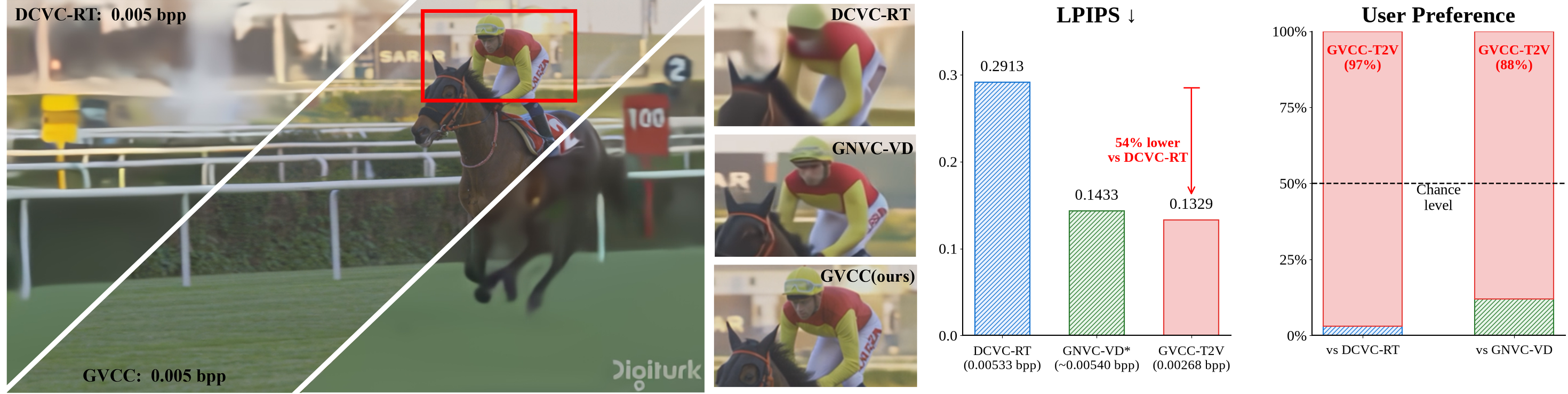}
\caption{%
\textbf{Qualitative, perceptual, and user-preference comparison.}
Left: the \textit{Jockey} example at the displayed sequence-specific operating points. Middle: full-UVG LPIPS-Alex; DCVC-RT (0.2913 at 0.00533\,bpp) and GVCC-T2V (0.1329 at 0.00268\,bpp) use our unified evaluation, while GNVC-VD$^*$ (0.1433 at ${\sim}$0.00540\,bpp) is read from its published curve. The displayed DCVC-RT and GVCC-T2V points differ by 54\%. Right: the user study preferred GVCC-T2V to DCVC-RT and GNVC-VD in 97\% and 88\% of comparisons, respectively.%
}
\label{fig:teaser}
\vspace{-10pt}
\end{figure}

This suggests a more direct formulation: rather than adding a generative refinement stage on top of a separately designed codec, we use the pretrained video generative model \emph{as} the codec itself, with the transmitted bitstream directly specifying the decoder's generative trajectory. We refer to this framework as \textbf{Generative Video Codebook Codec (GVCC)}.

A natural starting point is frame-wise compression with an image generative codec. However, temporal coherence is a property of the joint distribution across frames rather than of individual marginals, and independent per-frame generation cannot capture it: our preliminary experiments exhibit severe temporal flickering and appearance drift across GOPs. The compression backbone must therefore model the full video joint distribution, motivating the use of video foundation models.

Codebook-driven compression, following DDCM~\cite{ohayon2025ddcm}, NCS~\cite{su2025noiseneedsolvinglinear}, and Turbo-DDCM~\cite{vaisman2025turboddcm}, treats the decoder's per-step stochastic innovation as the channel through
which information is transmitted. Modern video generators such as Wan\,2.1~\cite{wan2025report} are built on rectified flow~\cite{liu2022rectified}, whose deterministic ODE sampler leaves no such channel once the initial latent is fixed. Crucially, the score-SDE framework~\cite{song2021scoreSDE} shows that a probability-flow ODE admits a family of reverse-time SDEs with the same continuous-time marginals. Introducing a nonzero diffusion coefficient can therefore be viewed as reallocating stochasticity from the initial latent to per-step innovations, thereby restoring the channel required by codebook-driven compression entirely at inference time, without retraining the generator.

Building on this unified SDE-codebook backbone, we instantiate three conditioning strategies occupying different points in the trade-off between spatial fidelity, temporal coherence, and bitrate. \textbf{T2V (Text-to-Video)} uses no reference frame, so the bitrate consists entirely of codebook indices, representing the lowest-side-information regime. \textbf{I2V (Image-to-Video)} uses a single reference frame with autoregressive GOP chaining to provide spatial anchoring across segments. \textbf{FLF2V (First-Last-Frame-to-Video)} uses both boundary frames of each GOP as dual anchors and constrains generation as interpolation across the segment.

\section{Related Work}\label{sec:related}

\paragraph{Generative Video Compression.}
Neural video codecs, building on learned image compression backbones~\cite{balle2018variational,minnen2018joint}, achieve strong rate--distortion performance~\cite{habibian2019video, hu2021fvc, sheng2022temporal, qi2025generative, jia2025dcvcrt}, but at ultra-low bitrates they often yield oversmoothed reconstructions with limited perceptual realism.
Recent methods address this issue by incorporating generative priors into the decoding pipeline.
GLVC~\cite{guo2025glvc} performs latent-space transform coding with spatio-temporal modeling, GNVC-VD~\cite{mao2025gnvcvd} uses a pretrained video diffusion transformer for sequence-level latent refinement, and GVC~\cite{chen2025gvc} shifts more reconstruction burden to receiver-side generative inference; these methods follow a hybrid paradigm where a separate codec defines the compressed representation and the generative model serves as a refinement module.
Free-GVC~\cite{ling2026freegvc} departs from this hybrid pattern by encoding directly along the diffusion trajectory via reverse channel coding, sharing the trajectory-based zero-shot philosophy of GVCC; the two differ in compression mechanism and backbone---reverse channel coding on CogVideoX-2B for Free-GVC, versus codebook-driven SDE sampling on rectified-flow video models for GVCC.

\paragraph{Zero-Shot Codebook-Driven Compression.}
DDCM, Turbo-DDCM and NCS show that a pretrained image diffusion model can be repurposed as a zero-shot image codec by replacing per-step Gaussian noise with reproducible codebook atoms and transmitting only compact indices.
However, these methods assume a stochastic reverse process with explicit noise injection at every step.
This assumption does not hold for modern rectified-flow video models such as Wan \, 2.1, whose sampling follows a deterministic ODE~\cite{liu2022rectified}.
Currently, \citet{kim2025inferencetime} also converts a rectified-flow ODE into an equivalent SDE at inference time, but does so to improve generation diversity and quality for inference scaling.
GVCC addresses this mismatch by converting rectified-flow sampling into an equivalent stochastic process at inference time, enabling codebook-driven compression for pretrained video generators without retraining.

\section{Method}\label{sec:method}

\begin{figure*}[t]
\centering
\includegraphics[width=\textwidth]{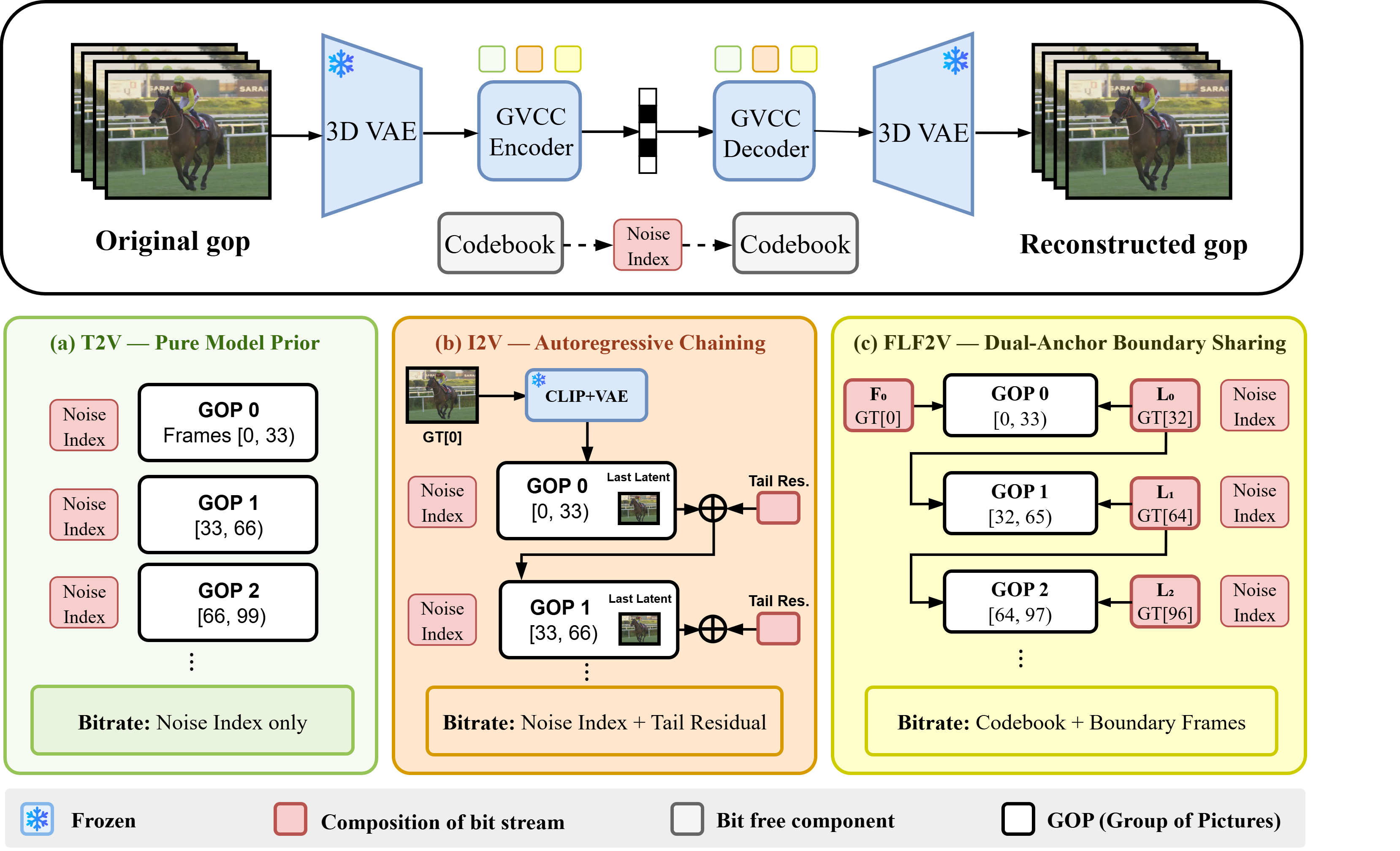}
\caption{%
\textbf{Overview of the GVCC framework.}
Top: shared pipeline---a frozen 3D VAE encodes the GOP into latent space, where GVCC compresses it into codebook noise indices; the decoder replays the same trajectory to reconstruct the video.
Bottom: three conditioning strategies.
(a)~T2V: codebook only, no reference frame.
(b)~I2V: autoregressive GOP chaining with tail residual correction.
(c)~FLF2V: dual-anchor boundary sharing across consecutive GOPs.}
\label{fig:framework}
\end{figure*}
As shown in Fig.\ref{fig:framework}, GVCC compresses each Group of Pictures (GOP) through a unified pipeline: a pretrained rectified-flow video model is converted into a stochastic process at inference time, and the per-step noise is replaced by reproducible codebook selections whose indices form the transmitted bitstream. All three conditioning variants---T2V,I2V and FLF2V---share the same SDE-codebook backbone and differ only in how the generative model is conditioned. We first review the stochastic-interpolant / flow-matching background and the RF specialization (Sec.~\ref{sec:bg_interp}), then describe ODE-to-SDE conversion (Sec.~\ref{sec:ode2sde}), score-from-velocity (Sec.~\ref{sec:score}), and codebook-driven discretization (Sec.~\ref{sec:codebook_sde}), before detailing the three conditioning strategies (Sec.~\ref{sec:t2v}--\ref{sec:flf2v}).

\subsection{Background: stochastic interpolants, flow matching, and rectified flow}\label{sec:bg_interp}

Diffusion models~\cite{ho2020ddpm,song2021ddim,karras2022edm}, flow matching, and rectified flow can all be viewed as learning dynamics along a family of probability paths $\{p_t\}_{0\le t\le 1}$ that interpolate between a data distribution $p_0$ and a noise distribution $p_1$~\cite{albergo2023unifying,lipman2022flow}.
A standard construction uses a \emph{stochastic interpolant} between two independent samples $\mathbf{x}_0\sim p_0$ and $\mathbf{x}_1\sim p_1$:
\begin{equation}\label{eq:interp_general}
  \mathbf{x}_t = \alpha_t\,\mathbf{x}_0 + \sigma_t\,\mathbf{x}_1, \quad t\in[0,1],
\end{equation}
where $\alpha_t$ and $\sigma_t$ are smooth scalars satisfying $\alpha_0{=}1$, $\sigma_0{=}0$, $\alpha_1{=}0$, $\sigma_1{=}1$, together with $\dot{\alpha}_t<0$ and $\dot{\sigma}_t>0$ so that $\mathbf{x}_t$ bridges $p_0$ and $p_1$~\cite{albergo2023building,albergo2023unifying}.
Different choices of $(\alpha_t,\sigma_t)$ induce different sampling trajectories; rectified flow~\cite{liu2022rectified} corresponds to the \emph{linear} choice
\begin{equation}\label{eq:rf_linear}
  \alpha_t = 1-t,\qquad \sigma_t = t,
\end{equation}
which yields straight paths in sample space and recovers the usual rectified-flow interpolation
\begin{equation}\label{eq:rf_interp}
  \mathbf{x}_t = (1-t)\,\mathbf{x}_0 + t\,\mathbf{x}_1, \qquad \mathbf{x}_1\sim\mathcal{N}(\mathbf{0},\mathbf{I}),
\end{equation}
when $p_1$ is isotropic Gaussian (equivalently, $\mathbf{x}_1\equiv\boldsymbol{\epsilon}$).
Flow matching~\cite{lipman2022flow} trains a velocity field $\mathbf{u}_t$ by regressing conditional vector fields along these paths; rectified-flow training then fits $\mathbf{u}_t$ to the instantaneous velocity of Eq.~\eqref{eq:rf_interp}. Sampling is usually performed by integrating the probability-flow ODE backward from $\mathbf{x}_1$ to $\mathbf{x}_0$:
\begin{equation}\label{eq:rf_ode}
  \mathrm{d}\mathbf{x}_t = \mathbf{u}_t(\mathbf{x}_t)\,\mathrm{d}t.
\end{equation}
This process is purely deterministic: given $\mathbf{x}_1$, the trajectory is unique, leaving no per-step degree of freedom for embedding codebook bits.

\subsection{ODE-to-SDE conversion}\label{sec:ode2sde}

To introduce controllable noise injection in Eq.~\eqref{eq:rf_ode}, we use the score-SDE construction~\cite{song2021scoreSDE}: for any probability-flow ODE generating $\{p_t\}$, there exists a family of reverse-time SDEs sharing the same marginals for any diffusion coefficient $g_t\ge 0$:
\begin{equation}\label{eq:reverse_sde}
  \mathrm{d}\mathbf{x}_t
  = \underbrace{\biggl[\mathbf{u}_t(\mathbf{x}_t) - \frac{g_t^2}{2}\,\nabla\!\log p_t(\mathbf{x}_t)\biggr]}_{\displaystyle \mathbf{f}_t(\mathbf{x}_t)}\mathrm{d}t
  \;+\; g_t\,\mathrm{d}\bar{\mathbf{w}},
\end{equation}

\paragraph{Diffusion schedule as a bit-budget constraint.}
Unlike prior ODE-to-SDE conversions motivated by sample diversity~\citep{kim2025inferencetime,ma2024sit}, the schedule $g_t$ in our setting is selected using a bit-budget consideration specific to codebook compression. At each Euler--Maruyama step, the stochastic innovation $g_t\sqrt{\Delta t}\,\mathbf{z}^*$ must be representable within the $M$-atom subspace spanned by the selected codebook entries. Under the near-orthogonality assumption of~\citet{vaisman2025turboddcm}, this subspace captures at most a fraction $c(M,d)$ of an arbitrary Gaussian innovation in latent dimension $d$, where $c(M,d)$ is monotonically increasing in $M$ and saturates as $M$ approaches the effective rank of the residual. In the video-latent setting, spatio-temporal correlation may weaken strict orthogonality; we therefore treat $c(M,d)$ as an empirical effective-rank quantity, with the saturation in Fig.~\ref{fig:ablation}(a) providing qualitative support rather than a direct estimate. The resulting design condition is
\begin{equation}
g_t^{2}\,\Delta t \;\le\; \rho \cdot c(M,d)\cdot \mathrm{Var}(r_t),
\label{eq:bit_budget}
\end{equation}
for some safety factor $\rho = \mathcal{O}(1)$, while also vanishing at $t=0$ to avoid spending bits on residuals below the codebook's resolution floor. We adopt
\begin{equation}
g_t = g_{\mathrm{scale}} \cdot t^2
\label{eq:g_schedule}
\end{equation}
as the simplest polynomial form satisfying both conditions. The scalar $g_{\mathrm{scale}}$ then serves as the primary rate--quality knob, subject to the upper bound in Eq.~\eqref{eq:bit_budget}. Appendix~\ref{app:param} verifies this
interpretation empirically: the narrow stability window $g_{\mathrm{scale}}\in[2,3]$ observed in Fig.~\ref{fig:ablation}(d) is 
consistent with the bound being saturated at $M=64$, and Fig.~\ref{fig:ablation}(a) confirms the corresponding saturation of $c(M,d)$ as $M$ grows.

\subsection{Score function from the velocity field}\label{sec:score}

The SDE drift in Eq.~\eqref{eq:reverse_sde} requires $\nabla\!\log p_t(\mathbf{x}_t)$, which the rectified-flow network does not output directly. For interpolants of the form Eq.~\eqref{eq:interp_general} with learned velocity $\mathbf{u}_t$, the score admits a closed form in terms of $(\alpha_t,\sigma_t,\dot{\alpha}_t,\dot{\sigma}_t)$~\cite{ma2024sit}:
\begin{equation}\label{eq:score_general}
  \nabla\!\log p_t(\mathbf{x}_t)
  = \frac{1}{\sigma_t}\cdot
  \frac{\alpha_t\,\mathbf{u}_t(\mathbf{x}_t) - \dot{\alpha}_t\,\mathbf{x}_t}
       {\dot{\alpha}_t\,\sigma_t - \alpha_t\,\dot{\sigma}_t}.
\end{equation}
Substituting the rectified-flow schedule [Eq.~\eqref{eq:rf_linear}] (i.e., $\alpha_t=1-t$, $\sigma_t=t$, $\dot{\alpha}_t=-1$, $\dot{\sigma}_t=1$) yields the RF specialization
\begin{equation}\label{eq:score_rf}
  \nabla\!\log p_t(\mathbf{x}_t)
  = -\,\frac{(1-t)\,\mathbf{u}_t(\mathbf{x}_t) + \mathbf{x}_t}{t}.
\end{equation}
Under Eq.~\eqref{eq:rf_interp}, $(1-t)\,\mathbf{u}_t + \mathbf{x}_t$ tracks the noise coordinate $\mathbf{x}_1$, recovering the usual score--noise relation. Eq.~\eqref{eq:score_rf} uses only the pretrained velocity network.

Substituting Eqs.~\eqref{eq:g_schedule} and~\eqref{eq:score_rf} into the drift of Eq.~\eqref{eq:reverse_sde} gives the explicit SDE drift
\begin{equation}\label{eq:sde_drift}
  \mathbf{f}_t(\mathbf{x}_t)
  = \mathbf{u}_t(\mathbf{x}_t)
  + \frac{g_t^2}{2}\cdot\frac{(1-t)\,\mathbf{u}_t(\mathbf{x}_t) + \mathbf{x}_t}{t}.
\end{equation}
When $g_t=0$, $\mathbf{f}_t$ reduces to $\mathbf{u}_t$, so the SDE strictly generalizes the RF ODE.

\subsection{Codebook-driven SDE sampling}\label{sec:codebook_sde}

We adopt the codebook mechanism of DDCM~\cite{ohayon2025ddcm} and Turbo-DDCM~\cite{vaisman2025turboddcm}, adapting it from DDPM reverse noise to the SDE [Eq.~\eqref{eq:reverse_sde}].
A single Euler--Maruyama (EM) step with step size $\Delta t$ takes the form
\begin{equation}\label{eq:em_step}
  \mathbf{x}_{t-\Delta t}
  = \mathbf{x}_t - \mathbf{f}_t(\mathbf{x}_t)\,\Delta t + g_t\sqrt{\Delta t}\;\zb,
  \quad \zb\sim\mathcal{N}(\mathbf{0},\mathbf{I}),
\end{equation}
where $\mathbf{z}$ is the per-step randomness to be replaced by codebook atoms. Following Turbo-DDCM's thresholding strategy, we replace it with a codebook-selected vector $\mathbf{z}^*$ as follows.

At each SDE step, the encoder first estimates the clean signal as $\hat{\mathbf{x}}_{0|t} = \mathbf{x}_t - t\cdot\mathbf{u}_t(\mathbf{x}_t)$ and computes the denoising residual $\mathbf{r} = \mathbf{x}_0 - \hat{\mathbf{x}}_{0|t}$, where $\mathbf{x}_0$ is the ground-truth latent available only at the encoder. A reproducible codebook $\mathcal{C}=\{\mathbf{z}^{(1)},\ldots,\mathbf{z}^{(K)}\}$ is then generated from a deterministic seed shared by both sides. The $M$ atoms with the largest absolute inner product with $\mathbf{r}$ are selected, along with their signs:
\begin{equation}\label{eq:atom_select}
  \{j_1,\ldots,j_M\} = \operatorname{top\text{-}M}_{i}\;\bigl|\langle\mathbf{z}^{(i)},\,\mathbf{r}\rangle\bigr|,
  \quad s_k = \mathrm{sign}\bigl(\langle\mathbf{z}^{(j_k)},\,\mathbf{r}\rangle\bigr).
\end{equation}
The selected atoms are combined and normalized to unit variance to match the theoretical noise magnitude of the SDE:
\begin{equation}\label{eq:composite}
  \zs = \frac{\sum_{k=1}^{M} s_k\,\mathbf{z}^{(j_k)}}
  {\mathrm{std}\!\bigl(\sum_{k=1}^{M} s_k\,\mathbf{z}^{(j_k)}\bigr)}.
\end{equation}
The noise in the codebook $\mathbf{z}^*$ then replaces the Gaussian innovation in the Euler update:
\begin{equation}\label{eq:sde_step}
  \mathbf{x}_{t-\Delta t}
  = \mathbf{x}_t - \mathbf{f}_t(\mathbf{x}_t)\,\Delta t + g_t\sqrt{\Delta t}\;\mathbf{z}^*.
\end{equation}
Since both encoder and decoder share the same seed, model weights, and codebook construction rule, transmitting only the $M$ indices and signs per step suffices for the decoder to reproduce the identical trajectory. For the last $N$ steps, we set $g_t=0$ and revert to the deterministic ODE $\mathbf{x}_{t-\Delta t} = \mathbf{x}_t - \mathbf{u}_t(\mathbf{x}_t)\,\Delta t$, which requires zero transmitted bits since both sides produce identical outputs from the synchronized preceding state.

\subsection{T2V: Pure Generative Prior Compression}\label{sec:t2v}

T2V corresponds to the conditioning-free regime of GVCC: no reference frame is transmitted, the model is driven only by an empty text prompt, and the entire bitrate is allocated to codebook indices.
This setting isolates the role of the pretrained video model as a spatio-temporal prior, with the codebook guiding an otherwise unconditioned generation trajectory toward the target video.

Because no spatial anchor is provided, the bitrate per GOP reduces to the codebook cost alone:
\begin{equation}\label{eq:bpp_t2v}
  \mathrm{BPP}_{\mathrm{T2V}} = \frac{(T{-}N)\cdot F \cdot B_{\mathrm{step}}}{F_{\mathrm{px}}\times H_{\mathrm{px}}\times W_{\mathrm{px}}},
\end{equation}
which yields the lowest bitrate among the three variants.
T2V therefore serves as a reference point for evaluating how much reconstruction quality can be obtained from the pretrained generative prior under codebook-only control.

A design trade-off of T2V is the lack of explicit spatial anchoring: without a reference frame, the decoded content may exhibit mild positional drift or appearance variation across GOPs.

\subsection{I2V: Autoregressive Compression with Tail Correction}\label{sec:i2v}

I2V conditions each GOP on a single reference frame, encoded through CLIP features and VAE latents, and therefore provides the strongest spatial anchoring among the three variants.
To reduce side-information cost, we use an autoregressive GOP structure: the first GOP takes a compressed version of the source's first frame as its reference, while each subsequent GOP reuses the decoded last frame of the previous GOP:
\begin{equation}\label{eq:i2v_ar}
  \text{GOP}_0\!:\; \mathrm{ref} = \tilde{\mathbf{I}}_0^{\mathrm{codec}}, \qquad
  \text{GOP}_{n>0}\!:\; \mathrm{ref} = \tilde{\mathbf{I}}_{n-1}^{\mathrm{last}}.
\end{equation}

We mitigate autoregressive error accumulation, which is concentrated on each GOP's last frame (both the hardest to reconstruct and the reference for the next GOP), with two complementary mechanisms.

First, \emph{adaptive tail-frame atom allocation} increases the codebook atom count from $M$ to $M_{\mathrm{tail}}$ for the last $F_{\mathrm{tail}}$ latent frames, allocating more bits to the temporally most difficult portion of the GOP.

Second, \emph{tail latent residual correction} transmits a quantized residual for the final latent frame.
After SDE encoding, the encoder computes the residual between the ground-truth latent and the decoded latent at the last temporal position, quantizes it to 8 bits per channel using min/max normalization, and compresses it losslessly with zlib.
At the decoder, this residual is added back to the reconstructed latent before VAE decoding, improving the quality of the frame that will be propagated as the next GOP reference.
The resulting overhead is included in the bitrate.

The total bitrate per GOP is:
\begin{equation}\label{eq:bpp_i2v}
  \mathrm{BPP}_{\mathrm{I2V}} = \frac{(T{-}N)\cdot F \cdot B_{\mathrm{step}}
  + B_{\mathrm{tail\_residual}} + \mathbb{1}_{n=0}B_{\mathrm{ref}}}
  {F_{\mathrm{px}}\times H_{\mathrm{px}}
  \times W_{\mathrm{px}}},
\end{equation}
where $B_{\mathrm{tail\_residual}}$ and $B_{\mathrm{ref}}$ denote the compressed
residual and initial-reference sizes in bits. The indicator charges the
serialized initial reference once; later references are inherited from decoded
outputs. All reported I2V bitrates include this initial-reference cost.

\subsection{FLF2V: Dual-Anchor Compression with Boundary Sharing} \label{sec:flf2v}

FLF2V conditions each GOP on both its first and last frames, providing dual temporal anchors that constrain the generative trajectory from both ends.
Compared with single-anchor I2V, this design is intended to limit temporal drift by requiring the model to satisfy boundary conditions at both the beginning and the end of the segment.

Both boundary frames are compressed with the \texttt{cheng2020-attn} image codec at quality level 4 in CompressAI~\cite{begaint2020compressai} and used to construct the conditioning signal.
Specifically, both frames are encoded through CLIP and placed at the temporal endpoints of the VAE latent volume, together with a binary mask indicating the conditioned positions.
The resulting generation process can be viewed as interpolation between two known endpoints, which is more constrained than extrapolation from a single anchor.

A key component of FLF2V is \emph{boundary-sharing GOP chaining}, which amortizes boundary-frame cost across consecutive GOPs.
The last frame of GOP $n$ is reused as the first frame of GOP $n{+}1$:
\begin{equation}\label{eq:flf2v_chain}
  \text{GOP}_n\!: \;\mathrm{first} = \tilde{I}_{n},\;\; \mathrm{last} = \tilde{I}_{n+1}, \qquad
  \text{GOP}_{n+1}\!: \;\mathrm{first} = \tilde{I}_{n+1},\;\; \mathrm{last} = \tilde{I}_{n+2}.
\end{equation}
Under this scheme, GOP\,0 transmits two boundary frames, while each subsequent GOP transmits only one new boundary frame, reducing boundary-frame overhead by approximately 50\% over long sequences.
The bitrate is:
\begin{equation}\label{eq:bpp_flf2v}
  \mathrm{BPP}_{\mathrm{FLF2V}} = \frac{(T{-}N)\cdot F \cdot B_{\mathrm{step}} + B_{\mathrm{boundary}}}{F_{\mathrm{px}}\times H_{\mathrm{px}}\times W_{\mathrm{px}}},
\end{equation}
where $B_{\mathrm{boundary}}$ equals $B_{\mathrm{first}} + B_{\mathrm{last}}$ (in bits) for the initial GOP and $B_{\mathrm{last}}$ for all subsequent GOPs.
Thus, $G$ consecutive GOPs contain $G+1$ paid anchor streams; neither the initial anchor nor the shared anchors are treated as free. Adjacent GOPs are conditioned on the same decoded boundary frame at their junction.

\section{Experiments}\label{sec:exp}

\subsection{Experimental Setup}\label{sec:setup}
All three GVCC variants use the Wan\,2.1 14B model family.
Unless otherwise specified, we evaluate on all seven UVG sequences~\cite{mercat2020uvg}. The main operating-point comparison uses 1080p; GVCC's native $1920{\times}1088$ output is resized, not cropped, to $1920{\times}1080$ for evaluation. The temporal stress test uses the \textit{Jockey} sequence at $1280{\times}720$. Each GVCC GOP contains 33 frames.
We additionally test a self-captured video to reduce concerns about possible overlap with the model's training distribution.

We report PSNR, MS-SSIM~\cite{wang2003msssim}, LPIPS~\cite{zhang2018lpips}, DISTS~\cite{ding2020dists}, and BPP. LPIPS uses the official AlexNet model (version 0.1), and DISTS uses the original implementation of Ding et al. All local reconstructions are evaluated with the same RGB pipeline and BT.601 color conversion. GNVC-VD has no public implementation; its approximate curve readings are therefore marked with an asterisk.

For a bitstream of $B$ bytes representing $F$ frames, we compute $\mathrm{BPP}=8B/(FHW)$ from the serialized stream, including transmitted reference frames and anchors. Evaluations use each codec's GOP-aligned output window; using the common first 94 aligned frames changes mean LPIPS and DISTS by at most 0.08\%. VMAF is evaluated on this common 94-frame window.

All 720p experiments are conducted on a single NVIDIA RTX 6000 Ada GPU with 48\,GB memory. Full hyperparameter settings are provided in Appendix~\ref{app:params}, and the default parameter selection procedure is described in Appendix~\ref{app:param}.

\subsection{Temporal Diagnostics Across GOPs}\label{sec:temporal}
Table~\ref{tab:temporal} reports a six-GOP, 720p stress test on the high-motion \textit{Jockey} sequence. Rates are computed from the sequence-specific serialized streams, including transmitted anchors.

At the displayed low-rate point, DCVC-RT has lower $E_{\mathrm{warp}}$, while T2V has lower tOF. Following Lai et al.~\cite{lai2018temporal}, $E_{\mathrm{warp}}$ uses a common RAFT-large~\cite{teed2020raft} backward flow estimated on each ground-truth pair, a forward--backward consistency mask, and RGB L1 warping error. tOF follows the optical-flow deviation of TecoGAN~\cite{chu2020tecogan}, and CLIP-F is adjacent-frame CLIP cosine similarity. Per-frame boundary LPIPS profiles are reported in Appendix~\ref{app:boundary_profiles}.

\subsection{Trade-offs Among GVCC Variants}\label{sec:variant_compare}

We first compare the three GVCC variants on all seven UVG sequences at
$1280{\times}720$, using three 33-frame GOPs per sequence. For reference, we
also evaluate DCVC-RT on the same resized sequences.

\begin{table*}[t]
\centering
\caption{\textbf{Compact temporal diagnostics on UVG \textit{Jockey} at 720p.} I2V's tail-residual stream is capped to a 0.050-bpp total budget.}
\label{tab:temporal}
\small
\setlength{\tabcolsep}{5pt}
\begin{tabular}{@{}llcccc@{}}
\toprule
Method & Setting & BPP & $E_{\mathrm{warp}}\downarrow$ & tOF$\downarrow$ & CLIP-F$\uparrow$ \\
\midrule
GVCC-T2V & --- & 0.0048 & 0.0102 & 0.565 & 0.9823 \\
GVCC-FLF2V & --- & 0.0096 & 0.0103 & 0.624 & 0.9838 \\
GVCC-I2V & capped & 0.050 & 0.0117 & 1.258 & 0.9830 \\
\midrule
GLC-Video & q0 & 0.0095 & 0.0179 & 2.949 & 0.9743 \\
DCVC-RT & q0 & 0.0045 & 0.0060 & 1.353 & 0.9812 \\
\bottomrule
\end{tabular}
\end{table*}

\begin{table*}[t]
\centering
\caption{%
\textbf{Comparison of GVCC variants on UVG 720p} (3 GOPs per sequence, 33 frames/GOP).
DCVC-RT at avg.\ ${\sim}$0.005\,bpp---matched to GVCC-T2V's bitrate---is included as a rate-matched perceptual reference.
Best and second-best results among GVCC variants are highlighted.%
}
\label{tab:variant}
\small
\setlength{\tabcolsep}{3pt}
\begin{tabular}{@{}l@{\hskip 6pt} ccc ccc ccc c@{}}
\toprule
 & \multicolumn{3}{c}{\textbf{GVCC-T2V}} & \multicolumn{3}{c}{\textbf{GVCC-I2V}} & \multicolumn{3}{c}{\textbf{GVCC-FLF2V}} & \textbf{DCVC-RT} \\
\cmidrule(lr){2-4} \cmidrule(lr){5-7} \cmidrule(lr){8-10} \cmidrule(l){11-11}
Seq. & PSNR$\uparrow$ & LPIPS$\downarrow$ & BPP & PSNR$\uparrow$ & LPIPS$\downarrow$ & BPP & PSNR$\uparrow$ & LPIPS$\downarrow$ & BPP & LPIPS$\downarrow$ \\
\midrule
Beau.  & 31.79 & 0.154 & 0.0048 & \colorbox{red!12}{32.90} & \colorbox{red!12}{0.109} & 0.0567 & \colorbox{blue!10}{32.26} & \colorbox{blue!10}{0.158} & 0.0075 & 0.442 \\
Bosp.  & 30.32 & 0.098 & 0.0048 & \colorbox{red!12}{33.63} & \colorbox{red!12}{0.055} & 0.0546 & \colorbox{blue!10}{32.72} & \colorbox{blue!10}{0.086} & 0.0108 & 0.233 \\
Honey. & 30.78 & 0.052 & 0.0048 & \colorbox{red!12}{36.21} & \colorbox{red!12}{0.020} & 0.0543 & \colorbox{blue!10}{34.73} & \colorbox{blue!10}{0.033} & 0.0141 & 0.180 \\
Jock.  & 31.28 & 0.090 & 0.0048 & \colorbox{red!12}{33.27} & \colorbox{red!12}{0.070} & 0.0534 & \colorbox{blue!10}{33.09} & \colorbox{blue!10}{0.080} & 0.0105 & 0.251 \\
RSG    & 26.74 & 0.086 & 0.0048 & \colorbox{red!12}{29.48} & \colorbox{red!12}{0.053} & 0.0516 & \colorbox{blue!10}{29.44} & \colorbox{blue!10}{0.063} & 0.0168 & 0.282 \\
SnD    & 25.55 & 0.239 & 0.0048 & \colorbox{red!12}{30.11} & \colorbox{red!12}{0.136} & 0.0549 & \colorbox{blue!10}{29.52} & \colorbox{blue!10}{0.166} & 0.0177 & 0.485 \\
Yacht. & 26.60 & 0.103 & 0.0048 & \colorbox{red!12}{28.21} & \colorbox{red!12}{0.075} & 0.0549 & \colorbox{blue!10}{28.02} & \colorbox{blue!10}{0.089} & 0.0141 & 0.315 \\
\midrule
\textbf{Avg.} & 29.01 & 0.117 & 0.0048 & \colorbox{red!12}{\textbf{31.97}} & \colorbox{red!12}{\textbf{0.074}} & 0.0543 & \colorbox{blue!10}{\textbf{31.40}} & \colorbox{blue!10}{\textbf{0.096}} & 0.0132 & 0.313 \\
\midrule
Codebook       & \multicolumn{3}{c}{18.4\,KB (100\%)} & \multicolumn{3}{c}{18.4\,KB ($\sim$9\%)}  & \multicolumn{3}{c}{18.4\,KB ($\sim$40\%)} & --- \\
Ref frames     & \multicolumn{3}{c}{0}                & \multicolumn{3}{c}{Compressed first + AR}      & \multicolumn{3}{c}{$\sim$28\,KB ($\sim$60\%)} & --- \\
Tail residual  & \multicolumn{3}{c}{N/A}              & \multicolumn{3}{c}{$\sim$189\,KB ($\sim$91\%)} & \multicolumn{3}{c}{N/A} & --- \\
\midrule
Spatial anchor    & \multicolumn{3}{c}{None}            & \multicolumn{3}{c}{Single (first)}    & \multicolumn{3}{c}{Dual (first + last)} & Learned \\
GOP structure     & \multicolumn{3}{c}{Independent}     & \multicolumn{3}{c}{AR chain}          & \multicolumn{3}{c}{Boundary sharing}    & P-frame \\
Strength          & \multicolumn{3}{c}{Lowest bitrate}  & \multicolumn{3}{c}{Highest fidelity}  & \multicolumn{3}{c}{Best balance}        & Best PSNR \\
\bottomrule
\end{tabular}
\end{table*}

Table~\ref{tab:variant} shows the original three-way trade-off. I2V achieves
the highest fidelity but spends substantially more bits on its tail residual;
T2V uses only codebook indices and occupies the lowest-rate regime; FLF2V lies
between them by sharing compressed boundary anchors.

\paragraph{Why three variants?}
The variants test three ways of addressing cross-GOP consistency: no explicit anchor (T2V), an autoregressive first-frame anchor (I2V), and shared endpoint anchors (FLF2V). I2V is an exploratory attempt to reduce discontinuities through autoregression; its present residual/codebook allocation is not yet balanced, as reflected by its bitrate and boundary profile (Appendix~\ref{app:boundary_profiles}).

\subsection{Rate--Distortion Curve Comparison}\label{sec:sota}

\begin{figure*}[t]
\centering
\includegraphics[width=\textwidth]{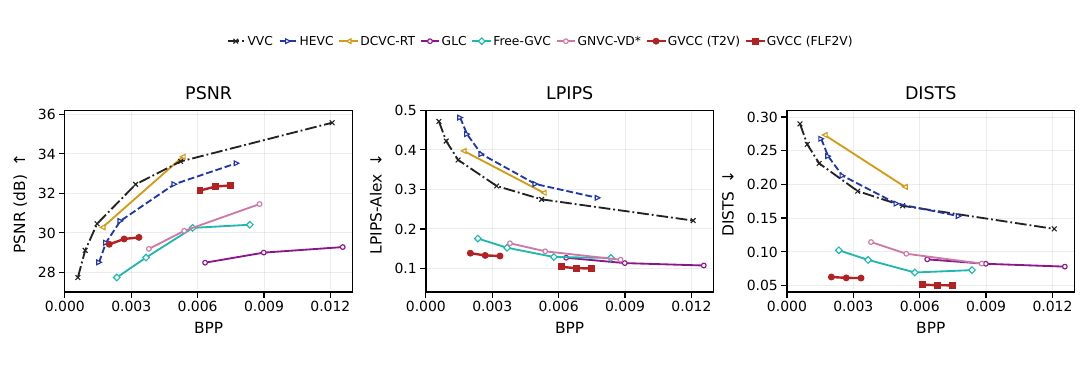}
\caption{%
\textbf{Full-UVG 1080p ultra-low-rate curves.}
From left to right: PSNR, LPIPS-Alex, and DISTS versus measured BPP. Only measured points below $0.013$\,bpp are shown, and lines connect operating points only within the same method or GVCC variant. I2V is outside this shared rate region and is reported in Appendix~\ref{tab:i2v_raw_points}. Free-GVC is our reproduction, and GNVC-VD$^*$ is digitized from its published curves.%
}
\label{fig:uvg_rd_curves}
\end{figure*}

For T2V and FLF2V, increasing $M$ moves each curve to the right while
improving all three reported quality metrics, with diminishing changes toward the
high-$M$ end. T2V occupies the lowest-rate region. FLF2V is shifted right by
its shared boundary anchors, while lying below T2V in LPIPS and DISTS and above
it in PSNR. These local relationships do not establish a global ordering among
the designs.

I2V lies in a separate rate region because its tail residual dominates the
stream. Its raw high-rate points are reported in Appendix~\ref{tab:i2v_raw_points}
rather than extrapolating baselines across a non-overlapping BPP interval.

In the overlapping low-rate regions, the T2V and FLF2V curves lie below the evaluated fidelity-oriented baselines in LPIPS and DISTS, while the latter generally retain higher PSNR. FLF2V also occupies a lower perceptual-distortion region than the evaluated generative baselines at nearby rates. The three GVCC variants are reported as distinct curves, without extrapolation across non-overlapping rate ranges.

HEVC and VVC use HM-16.25 and VTM-17.0, respectively; DCVC-RT and GLC-Video use released code and weights. Free-GVC is our implementation, while GNVC-VD is shown only as labeled readings from its published curves. Implementation details are listed in Appendix~\ref{app:implementations}.

Complementary VMAF, FVD, and spatial patch FID/KID results are provided in Appendix~\ref{app:additional_metrics}.

\section*{Limitations}
\textbf{Computational cost.}
GVCC relies on a 14B-parameter video DiT~\cite{peebles2023dit} with $T{=}20$ NFEs per GOP. On a single RTX 6000 Ada, encoding a 33-frame GOP takes ${\sim}90$\,s and decoding ${\sim}60$\,s, substantially slower than HEVC/VVC or learned predictive codecs~\cite{jia2025dcvcrt}. Improving runtime efficiency will likely require distillation, faster samplers, and hardware-aware optimization.

\textbf{Distribution shift.}
Faithfulness on controlled color/text/face tests and a self-captured out-of-distribution clip is reported in Appendix~\ref{app:faithful}.
On the OOD clip, perceptual quality (LPIPS 0.119) and color fidelity ($\Delta E=2.77$) remain in a usable range, while PSNR is several dB lower than on in-distribution sequences---a pattern consistent with the perception-oriented optimization of generative codecs (PSNR is not the design objective).
Behavior on content stylistically distant from common video distributions is therefore expected to vary, a property shared with other zero-shot generative methods rather than specific to GVCC.

\textbf{Rate curves and the I2V design.}
The variants occupy different rate ranges, and I2V's current tail correction remains bitrate-heavy; a lighter residual design is left for future work.

\textbf{Theoretical refinement under shifted-time schedules.}
Our diffusion coefficient $g_t = g_{\mathrm{scale}}\cdot t^2$ is formulated in the unshifted linear-time domain, whereas Wan\,2.1 uses an SD3-style~\cite{esser2024sd3} shifted reparameterization $\tau = st/(1+(s{-}1)t)$ with $s{=}5.0$.
While this formulation is empirically stable across our evaluations ($g_{\mathrm{scale}}{=}2.0$--$3.0$), extending the bit-budget analysis to SD3-style shifted reparameterizations is an interesting direction for tighter rate--quality control.

\section{Conclusion}
We presented GVCC, a zero-shot video compression framework that converts the deterministic ODE sampling of pretrained rectified-flow video models into codebook-driven stochastic decoding, with the diffusion coefficient formulated as a bit-budget-constrained design choice. T2V, I2V, and FLF2V expose different allocations of codebook, anchor, and residual bits. Their UVG rate--quality curves and temporal diagnostics demonstrate the potential of pretrained video generators as a foundation for zero-shot generative video compression.

{\small
\bibliographystyle{plainnat}
\bibliography{ref}
}


\appendix

\section{Conceptual Foundations of GVCC}\label{app:concepts}

Section~\ref{sec:method} derives GVCC formally; this appendix offers a
complementary conceptual reading of how the SDE+codebook construction operates
at bitrates down to approximately $0.003$\,bpp in our measured configuration.

\subsection{A coding channel via marginal-preserving SDE}\label{app:channel}
The rectified-flow ODE in Eq.~\eqref{eq:rf_ode} is fully deterministic, so
the only quantity through which a bitstream could influence reconstruction
is $\mathbf{x}_T$---prohibitively expensive to transmit at any nontrivial
precision. SDE conversion (Eq.~\eqref{eq:reverse_sde}) introduces a
per-step diffusion term whose marginals coincide with the ODE's by
construction~\citep{song2021scoreSDE}: it does not change \emph{what} the
model generates in expectation, only \emph{how} the trajectory is realized.
Each step now exposes a controllable perturbation through which the
bitstream can specify the reconstruction. Prior ODE-to-SDE
conversions~\citep{kim2025inferencetime,ma2024sit} use this construction to
widen sample diversity; GVCC repurposes the identical machinery to open a
coding channel, with $g_t$ acting as a bit-budget knob
(Eq.~\eqref{eq:bit_budget}). Although the SDE is mathematically stochastic,
in GVCC the executed trajectory is deterministic for both encoder and
decoder: the randomness in $g_t\,\mathrm{d}\bar{\mathbf{w}}$ is replaced by
codebook-selected innovations $\mathbf{z}^*$ uniquely determined by the
transmitted indices and the shared seed.

\subsection{Bits provide fidelity, prior provides plausibility}\label{app:duallabor}
The Euler--Maruyama update of Eq.~\eqref{eq:sde_step} decomposes each step
into two terms with complementary roles:
\begin{equation*}
\mathbf{x}_{t-\Delta t}
= \mathbf{x}_t \underbrace{- \mathbf{f}_t(\mathbf{x}_t)\,\Delta t}_{\text{drift (plausibility)}}
+ \underbrace{g_t\sqrt{\Delta t}\,\mathbf{z}^*}_{\text{noise (fidelity)}}.
\end{equation*}
The drift is \emph{source-agnostic}: computed from $\mathbf{u}_t$ and
$\mathbf{x}_t$ alone, with no access to $\mathbf{x}_0$, it keeps the
trajectory on the manifold of natural videos learned by the model. The
noise is \emph{source-driven}: the codebook-selected $\mathbf{z}^*$ is
constructed from the residual
$\mathbf{r}=\mathbf{x}_0-\hat{\mathbf{x}}_{0|t}$ and accumulates over the
$T-N$ noise-injection steps (cf.\ Sec.~\ref{sec:codebook_sde}) to steer the
trajectory toward $\mathbf{x}_0$. Neither term suffices
alone: pure drift yields a plausible video unrelated to $\mathbf{x}_0$;
pure noise escapes the manifold.

This division motivates the intended rate--perception behavior. Conventional
codecs must represent both visual structure and source-specific content,
whereas GVCC delegates part of the plausibility prior to the frozen generator
and uses transmitted indices to localize the reconstruction. Whether this
mechanism improves a rate--distortion or rate--perception curve must be judged
from matched multi-rate experiments; the present full-UVG results are reported
as measured BPP curves without a global BD claim.

\section{GOP-Boundary Profiles}\label{app:boundary_profiles}

\begin{figure*}[h]
\centering
\includegraphics[width=\textwidth]{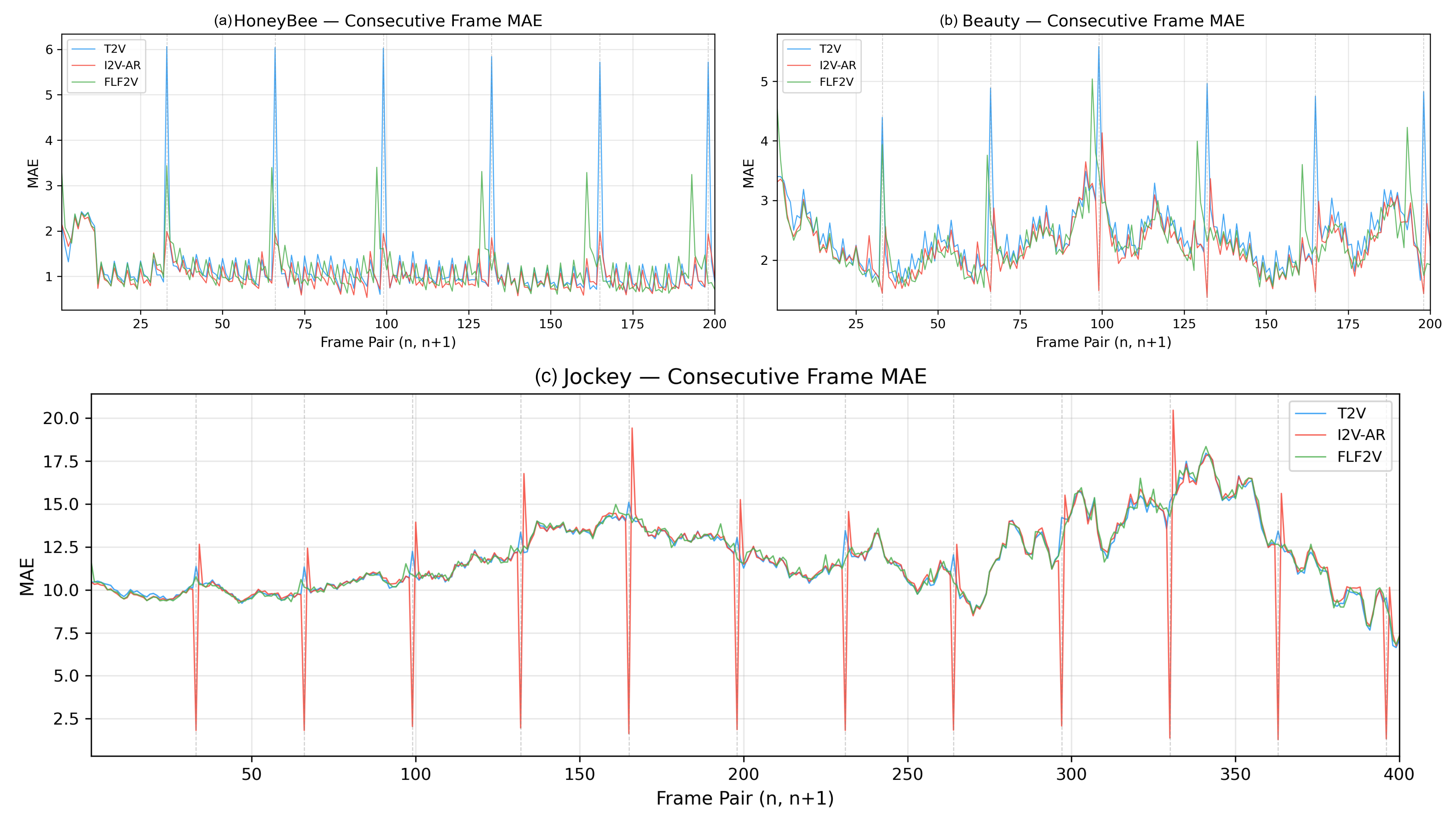}
\caption{Consecutive-frame MAE on \textit{HoneyBee}, \textit{Beauty}, and \textit{Jockey}. Dashed vertical lines indicate GOP boundaries.}
\label{fig:temporal_mae}
\end{figure*}

\paragraph{Consecutive-frame profiles.}
Figure~\ref{fig:temporal_mae} visualizes consecutive-frame MAE across GOPs. T2V shows periodic boundary spikes on \textit{HoneyBee} and \textit{Beauty}, while I2V shows sharp boundary changes on \textit{Jockey} associated with its tail correction.

\paragraph{Boundary LPIPS.}
We report per-frame LPIPS statistics around GOP boundaries. For each seam at output index $s$, the boundary set
contains frames $\{s-1,s\}$. ``Intra'' contains every other frame,
$\Delta_{\mathrm{boundary}}$ is boundary minus intra LPIPS, and the standard
deviation is computed over the complete per-frame LPIPS profile. The same
LPIPS-AlexNet protocol from Sec.~\ref{sec:setup} is used throughout. The seam
indices are $\{33,66,99,132,165\}$ for T2V/I2V and
$\{33,65,97,129,161\}$ for FLF2V.

\begin{table}[h]
\centering
\caption{Boundary LPIPS statistics on \textit{Jockey} at 720p. I2V uses the same capped stream as Table~\ref{tab:temporal}.}
\label{tab:boundary_lpips}
\small
\setlength{\tabcolsep}{3.5pt}
\begin{tabular}{@{}lccccc@{}}
\toprule
Variant & BPP & Frames / seams & Intra & Boundary & $\Delta$ / std. \\
\midrule
T2V   & 0.00483  & 198 / 5 & 0.0880 & 0.0898 & $+0.0018$ / 0.0048 \\
FLF2V & 0.009591 & 193 / 5 & 0.0780 & 0.0856 & $+0.0076$ / 0.0056 \\
I2V   & 0.050 (capped) & 198 / 5 & 0.0712 & 0.1147 & $+0.0435$ / 0.0217 \\
\bottomrule
\end{tabular}
\end{table}

\begin{figure*}[h]
\centering
\includegraphics[width=0.98\textwidth]{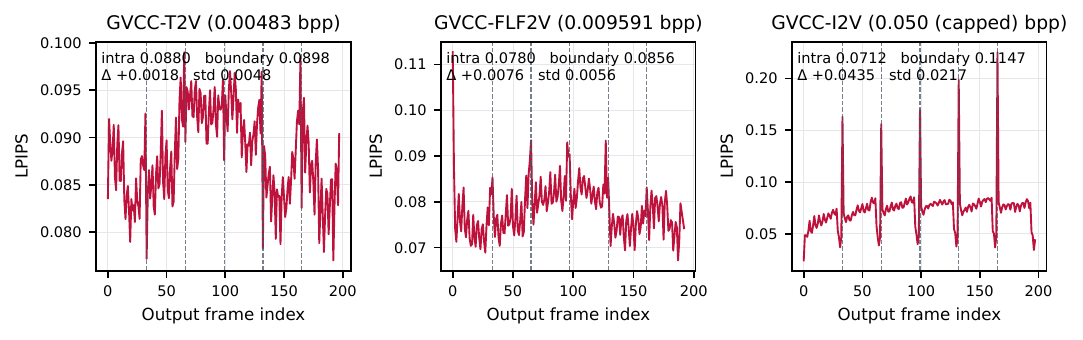}
\caption{Per-frame LPIPS profiles used for Table~\ref{tab:boundary_lpips}. Dashed lines mark GOP seam indices. The complete arrays, exact boundary indices, and evaluation code are included in the supplementary material.}
\label{fig:boundary_lpips}
\end{figure*}

\section{Codec and Metric Implementations}\label{app:implementations}

Table~\ref{tab:implementations} summarizes codec and metric implementations. GNVC-VD is the only curve-read result; the remaining plotted methods use local outputs and a unified metric pipeline.

\begin{table*}[h]
\centering
\caption{Codec and metric implementations used in evaluation.}
\label{tab:implementations}
\small
\setlength{\tabcolsep}{5pt}
\begin{tabular}{@{}p{0.18\textwidth}p{0.76\textwidth}@{}}
\toprule
Component & Version and configuration \\
\midrule
HEVC / VVC & HM-16.25 / VTM-17.0, low-delay configuration with intra period 32. \\
DCVC-RT / DCVC-FM & Official CVPR 2025 / 2024 weights; RT uses serialized streams and FM uses entropy-model rate estimates. \\
GLC-Video & Official released code and weights. \\
Free-GVC & Our implementation of the published method using CogVideoX-2B. \\
GNVC-VD & Approximate readings from the published RD curves; not locally recomputed. \\
LPIPS / DISTS & Official LPIPS-Alex v0.1 and the original DISTS implementation of Ding et al. \\
Additional metrics & VMAF v0.6.1, FVD with Kinetics-400 I3D, patch FID/KID with Inception-v3, RAFT-large optical flow, and CLIP ViT-B/32. \\
\bottomrule
\end{tabular}
\end{table*}

\section{Additional Perceptual Metrics and I2V Results}\label{app:additional_metrics}

\paragraph{VMAF.}
We evaluate the first 94 aligned frames of every UVG sequence at 1080p using
\texttt{vmaf\_v0.6.1} and a common BT.601 RGB-to-YUV420 conversion. Scores are
averaged per sequence and then equally across the seven sequences.

\begin{table*}[h]
\centering
\caption{VMAF on UVG 1080p. BPP denotes the measured operating point; no rate interpolation is applied.}
\label{tab:vmaf_appendix}
\footnotesize
\setlength{\tabcolsep}{3.5pt}
\begin{tabular}{@{}llcc@{\hskip 16pt}llcc@{}}
\toprule
Method & Setting & BPP & VMAF$\uparrow$ & Method & Setting & BPP & VMAF$\uparrow$ \\
\midrule
DCVC-RT & q42 & 0.01835 & 84.59 & GLC-Video & q42 & 0.01256 & 44.90 \\
VVC & QP32 & 0.01208 & 80.67 & GVCC-T2V & --- & 0.00268 & 44.25 \\
GVCC-FLF2V & --- & 0.00680 & 67.80 & GLC-Video & q21 & 0.00899 & 42.87 \\
GVCC-I2V & --- & 0.05230 & 67.59 & VVC & QP45 & 0.00146 & 42.02 \\
HEVC & QP37 & 0.00776 & 66.68 & Free-GVC & $t^*=250$ & 0.00578 & 41.83 \\
DCVC-RT & q21 & 0.00533 & 65.88 & HEVC & QP45 & 0.00251 & 40.51 \\
GLC-Video & q63 & 0.01644 & 46.10 & GLC-Video & q0 & 0.00634 & 39.79 \\
\bottomrule
\end{tabular}
\end{table*}

Despite its higher bitrate and lower LPIPS, I2V obtains nearly the same VMAF
as FLF2V (67.59 vs.\ 67.80). This suggests that its tail residual improves
perceptual detail without uniformly improving reference-aligned fidelity,
highlighting the current imbalance between residual and codebook bits.

\paragraph{FVD.}
We additionally report a descriptive FVD comparison on 233 identical,
non-overlapping 16-frame clips from all seven UVG sequences at 1080p. We use
the first 576 frames of each long sequence and the first 272 frames of
\textit{ShakeNDry}, a fixed 400-dimensional Kinetics-400 I3D representation,
and one Gaussian pooled over all clips. Every method is evaluated from
lossless reconstructions using the same clip indices. The operating points
are native and not rate matched; DCVC-RT uses serialized-stream BPP, GLC-Video
uses its released entropy-model estimate, and GVCC includes all transmitted
boundary anchors.

\begin{table}[h]
\centering
\caption{Descriptive FVD results on the fixed 233-clip UVG support. The
leave-one-sequence-out (LOO) range is reported as a sensitivity check. Since
the rates differ and the 400-dimensional covariance is rank deficient at this
sample count, no cross-method efficiency or distribution-level superiority is
inferred.}
\label{tab:fvd_appendix}
\footnotesize
\setlength{\tabcolsep}{4pt}
\begin{tabular}{@{}llccc@{}}
\toprule
Method & Setting & BPP & FVD$\downarrow$ & LOO range \\
\midrule
GVCC-FLF2V & $M=80$ & 0.006895 & 40.41 & 40.26--49.63 \\
GLC-Video & $q_I=3,q_P=63$ & 0.016982 & 41.64 & 41.25--51.43 \\
DCVC-RT & $q_I=q_P=63$ & 0.061345 & 9.69 & 4.87--11.80 \\
\bottomrule
\end{tabular}
\end{table}

\paragraph{Patch FID and KID.}
We evaluate 18,816 non-overlapping $256{\times}256$ patches from the first 96
frames of all seven UVG sequences using the same TorchMetrics 1.9.0
Inception-v3 pipeline. These metrics characterize spatial patch statistics and
do not measure temporal coherence.

\begin{table}[h]
\centering
\caption{Patch-level FID/KID on UVG 1080p under a unified local evaluation protocol.}
\label{tab:fid_kid_appendix}
\footnotesize
\setlength{\tabcolsep}{4pt}
\begin{tabular}{@{}llccc@{}}
\toprule
Method & Setting & BPP & FID$\downarrow$ & KID$\downarrow$ \\
\midrule
GVCC-FLF2V-14B & Local & 0.0070 & \textbf{14.92} & \textbf{0.00216} \\
Free-GVC & $t^*=215$ & 0.0072 & 18.36 & $0.00304 \pm 0.00031$ \\
\bottomrule
\end{tabular}
\end{table}

\paragraph{I2V high-rate operating points.}
\begin{table}[h]
\centering
\caption{Raw full-UVG 1080p I2V and DCVC-RT operating points. I2V BPP includes the initial reference frame.}
\label{tab:i2v_raw_points}
\footnotesize
\setlength{\tabcolsep}{2.5pt}
\begin{tabular}{llcccc}
\toprule
Method & Setting & BPP & PSNR $\uparrow$ & LPIPS $\downarrow$ & DISTS $\downarrow$ \\
\midrule
I2V & $M=60$  & 0.05163 & 32.72 & 0.0859 & 0.0501 \\
I2V & $M=80$  & 0.05230 & 32.83 & 0.0840 & 0.0497 \\
I2V & $M=100$ & 0.05297 & 32.86 & 0.0835 & 0.0496 \\
\midrule
DCVC-RT & q42 & 0.01835 & 36.94 & 0.2149 & 0.1420 \\
DCVC-RT & q63 & 0.06289 & 39.64 & 0.1661 & 0.1071 \\
\bottomrule
\end{tabular}
\end{table}

\paragraph{I2V residual budget.}
We vary the tail-residual budget on UVG \textit{Jockey} at 1080p while keeping
the remaining I2V configuration fixed.

\begin{table}[h]
\centering
\caption{I2V tail-residual budget sweep on \textit{Jockey} at 1080p.}
\label{tab:i2v_residual_sweep}
\footnotesize
\setlength{\tabcolsep}{4pt}
\begin{tabular}{@{}lcccl@{}}
\toprule
Residual budget & BPP & PSNR$\uparrow$ & LPIPS$\downarrow$ & Status \\
\midrule
Full & 0.050 & 32.83 & 0.084 & Stable \\
80\% & 0.041 & 30.12 & 0.108 & Stable \\
50\% & 0.027 & 27.53 & 0.172 & Stable \\
0\% & ${\sim}0.005$ & --- & --- & Unstable from GOP 3 \\
\bottomrule
\end{tabular}
\end{table}

The residual contribution is therefore not a smooth rate-control knob in the
current I2V design; I2V is treated as an exploratory variant.

\section{Hyperparameter Configuration}\label{app:params}

Table~\ref{tab:default} lists the full set of hyperparameters used across all three GVCC variants at 720p and 1080p resolutions.

\begin{table}[h]
\centering
\caption{Default hyperparameters for GVCC (720p / 1080p).}
\label{tab:default}
\small
\begin{tabular}{lccl}
\toprule
Parameter & 720p & 1080p & Role \\
\midrule
$M$ & 64 & 80 & Atoms per step (bitrate knob) \\
$M_{\mathrm{tail}}$ (I2V) & 128 & 128 & Atoms for AR tail frames \\
$K$ & 16384 & 16384 & Codebook size \\
$T$ & 20 & 20 & Total sampling steps \\
$N$ & 3 & 3 & Bit-free ODE tail steps \\
$g_{\mathrm{scale}}$ & 3.0 & 3.0 & SDE diffusion coefficient \\
GOP length & 33 & 33 & Frames per GOP ($4k{+}1$) \\
Ref quality (FLF2V) & 4 & 4 & CompressAI quality level \\
Tail residual (I2V) & 8-bit & 8-bit & Quantization for AR correction \\
Seed & 42 & 42 & Shared encoder/decoder seed \\
\bottomrule
\end{tabular}
\end{table}

\subsection{Default Parameter Selection}\label{app:param}

\begin{figure*}[t]
\centering
\includegraphics[width=0.85\textwidth]{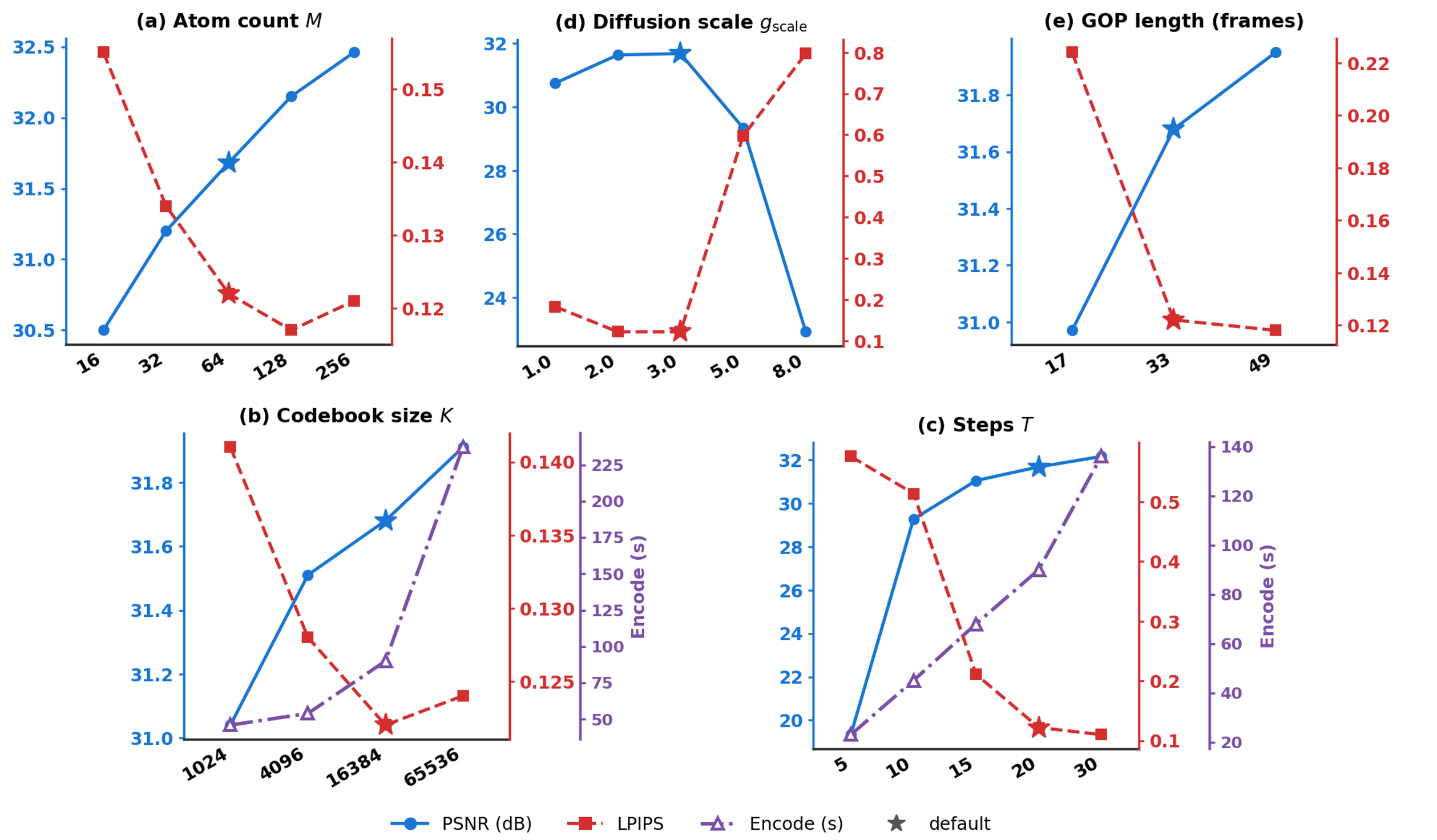}
\caption{%
\textbf{Hyperparameter sweeps on UVG \textit{Beauty} (T2V-1.3B, 720p).}
Blue: PSNR ($\uparrow$). Red: LPIPS ($\downarrow$). Purple: encoding time. Stars: selected defaults.
(a)~Atom count $M$: quality saturates around $M{=}64$ while BPP grows linearly.
(b)~Codebook size $K$: diminishing returns beyond 16384 at rapidly increasing cost.
(c)~Steps $T$: marked degradation at $T{=}5$ and smaller changes beyond $T{=}20$.
(d)~Diffusion scale $g_{\mathrm{scale}}$: narrow sweet spot at 2.0--3.0; collapse at higher values.
(e)~GOP length: 17 frames insufficient; 33 and 49 comparable.%
}
\label{fig:ablation}
\end{figure*}

We sweep each hyperparameter individually on the UVG \textit{Beauty} sequence using T2V-1.3B at 720p, varying one parameter at a time while holding the others at default values. Results are shown in Fig.~\ref{fig:ablation}.

\paragraph{Atom count $M$ (Fig.~\ref{fig:ablation}a).}
$M$ is the primary bitrate control variable, with BPP scaling nearly linearly from 0.0012 ($M{=}16$) to 0.0192 ($M{=}256$). Increasing $M$ from 16 to 64 raises PSNR by 1.2\,dB at 0.0048\,BPP. Beyond $M{=}128$, the measured change is smaller---$M{=}256$ adds 0.3\,dB while LPIPS changes from 0.117 to 0.121. Encoding time is approximately 91\,s across these settings. We use $M{=}64$ as the default in subsequent experiments.

\paragraph{Codebook size $K$ (Fig.~\ref{fig:ablation}b).}
$K$ determines the size of the per-step search space. Increasing $K$ from 1024 to 16384 changes PSNR from 31.04 to 31.68\,dB and LPIPS from 0.141 to 0.122, while encoding time changes from 46 to 90\,s. Increasing to $K{=}65536$ gives 0.23\,dB additional PSNR at 237\,s. We therefore use $K{=}16384$ for the reported experiments.

\paragraph{Sampling steps $T$ (Fig.~\ref{fig:ablation}c).}
The number of SDE steps has a large measured effect. At $T{=}5$, PSNR is 19.3\,dB and LPIPS is 0.576; LPIPS changes from 0.514 to 0.211 between $T{=}10$ and $T{=}15$. From $T{=}20$ to $T{=}30$, PSNR increases by 0.5\,dB while encoding time changes from 90 to 136\,s. We use $T{=}20$ as the reported default.

\paragraph{Diffusion scale $g_{\mathrm{scale}}$ (Fig.~\ref{fig:ablation}d).} This parameter exhibits a notably narrow effective range. $g_{\mathrm{scale}}{=}2.0$ and $3.0$ perform nearly identically (PSNR 31.64 vs. 31.68, LPIPS 0.122 vs. 0.122). However, quality collapses rapidly outside this window: $g_{\mathrm{scale}}{=}5.0$ 
drops PSNR to 29.3 dB with LPIPS 0.597, and $g_{\mathrm{scale}}{=}8.0$ gives 22.9 dB / 0.797. This behavior is consistent with the bit-budget interpretation of Eq.~\eqref{eq:bit_budget}: increasing $g_{\mathrm{scale}}$ scales the innovation variance $g_t^2\,\Delta t$ quadratically, and the fixed $M{=}64$ atom set eventually becomes insufficient for residual matching. The atom-count sweep in Fig.~\ref{fig:ablation}(a) likewise shows diminishing changes beyond $M{=}128$. We use $g_{\mathrm{scale}}{=}3.0$ in the reported configuration.

\paragraph{GOP length (Fig.~\ref{fig:ablation}e).}
A GOP of 17 frames is too short for the video model to fully exploit its temporal modeling capacity, resulting in substantially worse perceptual quality (LPIPS 0.224 vs.\ 0.122 at 33 frames). Extending to 49 frames yields marginal improvement (LPIPS 0.118, PSNR +0.27\,dB) at slightly lower BPP (0.00471 vs.\ 0.00483), since the VAE's $4{\times}$ temporal stride compresses longer sequences more efficiently. We adopt 33 frames as the default, noting that 49 frames is a viable option for 1080p where the additional compute is justified.

\paragraph{Summary.}
The validated default configuration is $M{=}64$, $K{=}16384$, $T{=}20$, $g_{\mathrm{scale}}{=}3.0$, GOP${=}33$. The complete parameter table with 720p and 1080p settings is provided in Table~\ref{tab:default}. These sweeps are mainly used to determine stable and interpretable default trends, and a sanity check has also been done on the main model.

\section{Faithfulness and Robustness}\label{app:faithful}

A central concern for generative codecs is whether they reconstruct the source
faithfully or merely produce plausible alternatives. To evaluate this, we
conduct four targeted tests using GVCC-FLF2V at 720p (3 GOPs, 97 frames per
video), focusing on color fidelity, text fidelity, face identity, and
out-of-distribution (OOD) behavior. Results are summarized in
Table~\ref{tab:faithful}.

\begin{table}[h]
\centering
\caption{%
\textbf{Faithfulness and robustness evaluation} (GVCC-FLF2V, 720p, 3 GOPs).
$\Delta E$: CIEDE2000 color difference ($<$1: imperceptible, 1--3: barely
noticeable, $>$3: clearly visible). Face ID: ArcFace~\cite{deng2019arcface} cosine similarity between
original and reconstructed faces.
}
\label{tab:faithful}
\small
\setlength{\tabcolsep}{3pt}
\begin{tabular}{@{}l@{\hskip 4pt} ccccc@{}}
\toprule
Test & PSNR$\uparrow$ & LPIPS$\downarrow$ & MS-SSIM$\uparrow$ & $\Delta E$$\downarrow$ & Face ID$\uparrow$ \\
\midrule
Color & 48.60 & 0.004 & 0.998 & 0.26 & --- \\
Text  & 38.87 & 0.002 & 0.998 & 0.26 & --- \\
Face  & 31.58 & 0.022 & 0.986 & 2.34 & 0.857 \\
OOD   & 25.15 & 0.119 & 0.910 & 2.77 & --- \\
\bottomrule
\end{tabular}
\end{table}

\textbf{Controlled fidelity tests.}
The color-chart and text-overlay tests yield $\Delta E=0.26$ in both cases,
indicating imperceptible color deviation. For text, GVCC additionally achieves
PSNR 38.87\,dB and LPIPS 0.002, indicating faithful reconstruction of
high-contrast symbolic content under controlled in-distribution conditions.

\textbf{Face identity.}
We measure ArcFace cosine similarity on 8 sampled frames and obtain a mean of
0.857 ($\sigma=0.054$, range 0.765--0.937), with all sampled frames preserving
the correct number of detected faces. This indicates moderate but not severe
identity degradation. GOP-level PSNR remains stable across the chain
(31.1$\to$32.0$\to$31.7\,dB) and $\Delta E$ stays at 2.34, suggesting no clear
evidence of cumulative identity drift across the three chained GOPs.

\textbf{Out-of-distribution behavior.}
We additionally evaluate a self-captured clip center-cropped and resized to
720p. Perceptual quality and color fidelity remain in a usable range
(LPIPS 0.119, MS-SSIM 0.910, $\Delta E=2.77$), while PSNR is several dB lower
than on the controlled in-distribution tests; this gap is consistent with the
perception-oriented optimization of generative codecs and is more pronounced
in PSNR than in perceptual or color metrics. Boundary-frame overhead rises
modestly to about 66\% of the per-GOP bitstream (108/163\,KB), versus
roughly 60\% on the in-distribution setting, reflecting content-driven
entropy variation in the learned image codec.

\textbf{Summary.}
Across all four tests, per-GOP PSNR remains within roughly $\pm$1.5\,dB
across the chain (e.g., 25.9$\to$24.9$\to$24.6\,dB on the OOD clip;
31.1$\to$32.0$\to$31.7\,dB on the face test), with no monotonic decay---
consistent with the boundary-sharing design of FLF2V
(Sec.~\ref{sec:flf2v}).

\subsection{VAE Upper Bound Analysis}\label{app:vae}

To disentangle the quality loss introduced by the 3D VAE from that of the codebook compression, we compare the VAE encode--decode upper bound (no GVCC, just VAE round-trip) against the full FLF2V pipeline on the four faithfulness test videos. Results are shown in Table~\ref{tab:vae_bound}.

\begin{table}[h]
\centering
\caption{VAE upper bound vs.\ FLF2V compression (720p).}
\label{tab:vae_bound}
\small
\begin{tabular}{@{}lccc@{}}
\toprule
Test & VAE PSNR & FLF2V PSNR & $\Delta$ (GVCC loss) \\
\midrule
Color     & 52.63 & 48.60 & $-$4.03 \\
Text      & 40.46 & 38.87 & $-$1.59 \\
Face      & 35.01 & 31.58 & $-$3.43 \\
OOD       & 30.86 & 25.15 & $-$5.71 \\
\bottomrule
\end{tabular}
\end{table}

The additional loss from GVCC beyond the VAE upper bound is small on the controlled in-distribution tests (text: $-$1.59\,dB; color: $-$4.03\,dB)---the VAE itself retains 40--52\,dB PSNR---suggesting that the dominant degradation comes from the codebook-driven compression stage rather than from the VAE alone. For the face test, the VAE retains very high identity similarity (cosine similarity 0.987), while the full pipeline drops to 0.857, indicating that the remaining identity shift is primarily associated with the codebook-driven generative reconstruction process. The largest gap appears on the out-of-distribution clip ($-$5.71\,dB). Notably, the VAE round-trip on this clip ($30.86$\,dB) is markedly lower than on the controlled in-distribution tests ($40{+}$\,dB), indicating that a substantial portion of the OOD degradation originates from latent reconstruction rather than from the codebook stage alone. Since the codebook approximation error scales inversely with $M$, increasing the atom count is one possible direction for further narrowing this remaining gap, subject to the saturation of $c(M,d)$ observed in Fig.~\ref{fig:ablation}(a).

\section{Rate--Distortion Sweep}\label{app:rd_sweep}

\begin{figure}[h]
\centering
\includegraphics[width=0.56\columnwidth]{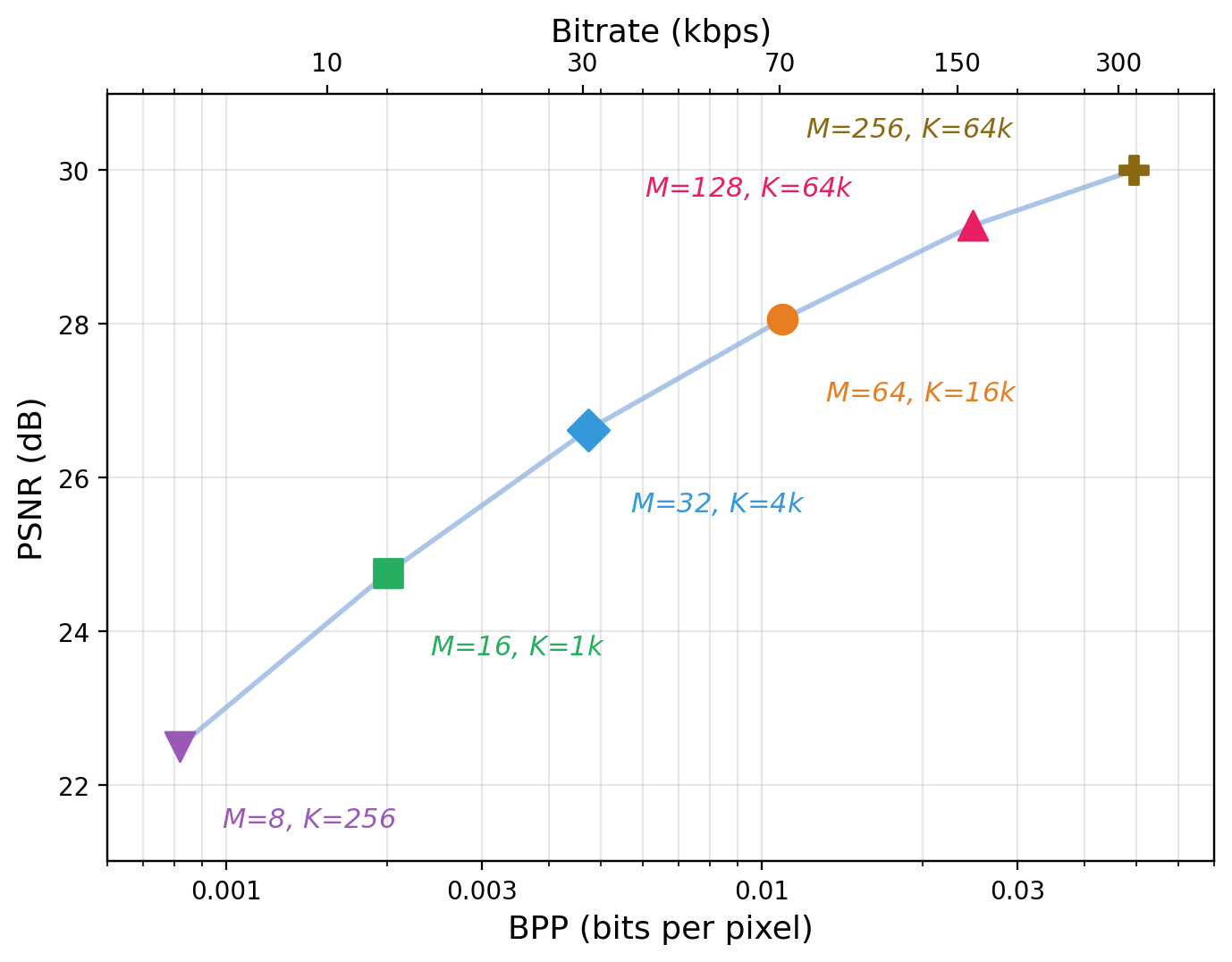}
\caption{%
\textbf{Rate--distortion curve of GVCC-T2V} (1.3B, 480p, UVG average).
Each point corresponds to a $(M, K)$ configuration from Table~\ref{tab:rd_sweep}. The curve spans from 5.3\,kbps to 322.8\,kbps with monotonically increasing quality.%
}
\label{fig:rd_sweep}
\end{figure}

To characterize GVCC's rate--distortion behavior over a wide operating range, we sweep 11 $(M, K)$ configurations using T2V-1.3B at 480p (7 UVG sequences, 3 GOPs each). Results are shown in Table~\ref{tab:rd_sweep} and Fig.~\ref{fig:rd_sweep}. BPP and PSNR increase monotonically from 0.0008\,bpp / 22.5\,dB to 0.0496\,bpp / 30.0\,dB, confirming smooth and predictable bitrate control across nearly two orders of magnitude.

\begin{table}[h]
\centering
\caption{T2V-1.3B R-D sweep on UVG 480p (7 seq.\ $\times$ 3 GOPs).}
\label{tab:rd_sweep}
\small
\begin{tabular}{@{}rrcrr@{}}
\toprule
$M$ & $K$ & PSNR (dB) & BPP & kbps \\
\midrule
8   & 256   & 22.50 & 0.00081 & 5.3 \\
16  & 512   & 24.32 & 0.00183 & 11.9 \\
16  & 1024  & 24.76 & 0.00201 & 13.1 \\
32  & 2048  & 26.32 & 0.00438 & 28.5 \\
32  & 4096  & 26.62 & 0.00474 & 30.9 \\
64  & 4096  & 27.61 & 0.00948 & 61.7 \\
64  & 16384 & 28.06 & 0.01095 & 71.2 \\
128 & 16384 & 28.94 & 0.02187 & 142.4 \\
128 & 65536 & 29.28 & 0.02478 & 161.4 \\
256 & 16384 & 29.64 & 0.04374 & 284.9 \\
256 & 65536 & 30.00 & 0.04956 & 322.8 \\
\bottomrule
\end{tabular}
\end{table}

\section{Broader Impacts}
By reducing bandwidth and storage costs, GVCC may benefit low-bandwidth deployments and energy-efficient video infrastructure. As a generative codec, it can hallucinate plausible but inaccurate content at extreme bitrates (Appendix~\ref{app:faithful}), making it unsuitable for forensic or evidentiary use; outputs should be flagged as generative reconstructions in such contexts.

\end{document}